\documentclass[conference]{IEEEtran}
\IEEEoverridecommandlockouts
\usepackage{cite}
\usepackage{amsmath,amssymb,amsfonts}
\usepackage{multirow}
\usepackage{booktabs}
\usepackage{url}
\usepackage{graphicx}
\usepackage{balance}
\usepackage{textcomp}
\usepackage{xcolor}
\def\BibTeX{{\rm B\kern-.05em{\sc i\kern-.025em b}\kern-.08em
    T\kern-.1667em\lower.7ex\hbox{E}\kern-.125emX}}
    
\newif\ifanonymous
\anonymousfalse 
\begin{document}

\title{NeuroECG: ECGFounder-Based Deep ECG Representation for EEG-Free Neurological Prognostication After Cardiac Arrest\\
}
\ifanonymous
\author{\IEEEauthorblockN{Anonymous Author(s)}
  }
\else
\author{

\IEEEauthorblockN{
1\textsuperscript{st} Jiajun Gao
}
\IEEEauthorblockA{
\textit{School of Economics and Management} \\
\textit{Xidian University} \\
Xi'an, China \\
25069100156@stu.xidian.edu.cn
}

\and

\IEEEauthorblockN{
2\textsuperscript{nd} Yi Zhao
}
\IEEEauthorblockA{
\textit{School of Mechano-Electronic Engineering} \\
\textit{Xidian University} \\
Xi'an, China \\
24049200262@stu.xidian.edu.cn
}

\and

\IEEEauthorblockN{
3\textsuperscript{rd} Chenyang Xu
}
\IEEEauthorblockA{
\textit{School of Cyber Engineering} \\
\textit{Xidian University} \\
Xi'an, China \\
xcy@ieee.org
}

\and

\IEEEauthorblockN{
4\textsuperscript{th} Yuxi Zhou\textsuperscript{*}
}
\IEEEauthorblockA{
\textit{Department of Computer Science} \\
\textit{Tianjin University of Technology} \\
Tianjin, China \\
\textit{DCST, BNRist, RIIT, Institute of Internet Industry} \\
\textit{Tsinghua University} \\
Beijing, China \\
joy\_yuxi@pku.edu.cn
}

\and

\IEEEauthorblockN{
5\textsuperscript{th} Hao Wang\textsuperscript{*}
}
\IEEEauthorblockA{
\textit{School of Cyber Engineering} \\
\textit{Xidian University} \\
Xi'an, China \\
Haow@ieee.org
}

\thanks{This work was supported by “the Fundamental Research Funds for the Central Universities” under Grant X202610701781.}
\thanks{
Hao Wang\textsuperscript{*} and Yuxi Zhou\textsuperscript{*}
are the corresponding authors
(e-mail: Haow@ieee.org; joy\_yuxi@pku.edu.cn).
}

}

\fi
\maketitle

\begin{abstract}
Neurological prognostication after cardiac arrest commonly relies on electroencephalography (EEG). However, EEG demands high clinical resources. Bedside electrocardiography (ECG) is standard and low-cost. Yet, its value for predicting neurological outcomes remains underexplored. 
In this study, we propose NeuroECG, an ECGFounder-based deep representation framework for EEG-free auxiliary prognostication. 
NeuroECG adapts a pretrained ECG foundation model via task-specific fine-tuning. We implement a gradual unfreezing strategy on single-channel bedside monitoring ECG. 
Multiple ECG segments per patient are encoded into segment-level deep features. These embeddings are aggregated via quantile pooling ($q = 0.24$) and compressed using principal component analysis (PCA). 
Experiments on 412 ECG-available patients from the multicenter I-CARE database show that the adapted ECGFounder backbone achieves the best ECG-only AUROC among evaluated backbones (0.7333). Fusion with static clinical covariates reaches a test AUROC of 0.8077 and AUPRC of 0.8970, above the static-only point estimates. The paired-bootstrap confidence interval for the AUROC difference includes zero. These findings suggest that bedside ECG representations may provide auxiliary prognostic information in an EEG-free setting.
The source code is available at \url{https://github.com/goddream66/NeuroECG}.
\end{abstract} 


\section{Introduction}
Neurological prognostication after cardiac arrest is a critical task in intensive care medicine. Early assessment of neurological recovery is crucial for risk stratification and clinical decision-making. Standard guidelines recommend multimodal assessments, with EEG as a cornerstone modality~\cite{nolan2021european}. Recent deep learning studies have also shown that EEG dynamics contain strong prognostic information in comatose patients after cardiac arrest~\cite{zheng2022predicting}. However, continuous EEG requires specialized hardware, trained technicians, and expert interpretation~\cite{bongiovanni2020standardized}. These constraints limit its availability in resource-limited or emergency settings~\cite{chen2024electroencephalogram}. In contrast, bedside monitoring ECG is routinely acquired in ICU patients at low cost. ECG does not directly measure cortical activity. Yet, post-resuscitation pathophysiology can alter cardiac morphology and rhythm dynamics through the heart--brain axis~\cite{tahsili2017heart}. Therefore, bedside ECG may provide an accessible auxiliary signal for EEG-free prognostication.

However, prior ECG-based prognostic methods remain limited in this setting. Conventional ECG descriptors are sensitive to noise and recording artifacts. They often show weak standalone predictive power compared with EEG or clinical variables~\cite{niu2025explainable}. These limitations motivate representation learning directly from raw bedside ECG. Nevertheless, training complex deep models from scratch is difficult on clinical datasets. Post-cardiac-arrest cohorts are usually small, heterogeneous, and imbalanced. High-capacity networks can easily overfit under these conditions~\cite{zhou2023less}.

Pretrained ECG foundation models have emerged as a compelling way to mitigate data scarcity~\cite{li2025electrocardiogram,mckeen2025ecg}. These models learn generalizable morphology representations from large ECG datasets. They can serve as robust feature extractors for downstream clinical tasks. However, their use in ICU bedside monitoring remains underexplored. Continuous bedside ECG produces massive multi-segment recordings. These segment-level embeddings must be converted into fixed-dimensional patient-level representations. This step is challenging. It is further complicated by multicenter monitoring inconsistencies. Examples include heterogeneous channel names and variable recording durations in the I-CARE registry~\cite{icare2023}.

To address these challenges, we propose NeuroECG, a deep ECG representation framework for EEG-free auxiliary neurological prognostication after cardiac arrest. The framework is designed for continuous, raw bedside ICU monitoring records from the I-CARE database~\cite{icare2023}. As shown in Fig.~\ref{fig:framework}, NeuroECG fine-tunes a pretrained ECG foundation model, ECGFounder, using gradual unfreezing. This strategy helps maintain backbone stability under limited clinical data. The framework encodes 10-second bedside ECG segments into deep embeddings. These embeddings are aggregated to the patient level by feature-wise quantile pooling ($q = 0.24$). They are then compressed using PCA. Finally, the compact ECG representation is combined with static clinical covariates. A CatBoost classifier~\cite{prokhorenkova2018catboost} is used for binary outcome prediction. A CatBoost regressor is used for auxiliary functional score estimation. The main contributions of this study are summarized as follows:
\begin{itemize}
    \item We establish a standardized preprocessing and flatline-oriented quality-control workflow for continuous bedside ICU ECG in the I-CARE database~\cite{icare2023}. This provides a practical benchmark for future waveform studies.
    \item We adapt a pretrained ECG foundation model to single-channel bedside monitoring ECG via gradual unfreezing. This forms a task-specific transfer learning workflow for low-cost, EEG-free auxiliary prognostication.
    \item We construct a lightweight patient-level representation pipeline. It aggregates continuous ICU recordings into deep ECG features using quantile pooling and PCA-based compression.
    \item We combine the learned deep ECG representations with static clinical covariates. This evaluates their complementary prognostic value over strong clinical baselines.
\end{itemize}

\section{Related Work}

\subsection{ECG-Based Prognostication After Cardiac Arrest}

ECG is widely used for rhythm monitoring and arrhythmia detection~\cite{hannun2019cardiologist}. After cardiac arrest, ECG changes may also reflect systemic and neurocardiac injury through the heart--brain axis~\cite{tahsili2017heart}. Recent work has explored this prognostic potential. Takahashi et al.~\cite{takahashi2023electrocardiogram} analyzed lead-II ICU monitoring ECG for neurological outcome prediction after out-of-hospital cardiac arrest. Their framework derived 71 diagnostic statement probabilities using a PTB-XL-based premodel and incorporated them into ECG outcome modeling. The best AUROC for favorable neurological outcome was 0.688. This study supports the prognostic value of routinely collected monitoring ECG. However, continuous ICU ECG is noisy and often contains missing signals or electrode-off flatlines~\cite{icare2023}. Direct deep representation learning from raw bedside ECG also remains underexplored. These gaps motivate more robust preprocessing and representation learning for continuous monitoring ECG.

\subsection{Pretrained ECG Foundation Models}

Pretrained ECG foundation models provide a promising approach to ECG representation learning. ECG-FM uses masked and contrastive pretraining on large ECG databases~\cite{mckeen2025ecg}. ECG-JEPA learns general ECG representations through joint-embedding predictive learning~\cite{kim2024learning}. ST-MEM uses spatio-temporal masked reconstruction to capture ECG dynamics~\cite{na2024guiding}. Other models further connect ECG signals with clinical knowledge or language. GEM combines time-series and image inputs for grounded ECG understanding~\cite{lan2025gemempoweringmllmgrounded}. Zero-shot ECG models use clinical knowledge during inference~\cite{liu2024zeroshotecgclassificationmultimodal}. TokenRhythm learns multi-scale rhythm representations~\cite{wang2025tokenrhythmmultiscaleapproach}. Similar foundation-model ideas have also been explored for other physiological signals, such as neural interfaces~\cite{NEURIPS2025_dd9c5ce8}. These studies motivate the use of pretrained ECG representations for bedside ECG outcome prediction.

\subsection{Transfer Learning and Multimodal ICU Prediction}

Clinical cohorts are often small and heterogeneous. Transfer learning can improve generalization under these conditions~\cite{raghu2019transfusion}. Gradual unfreezing and discriminative learning rates can also reduce catastrophic forgetting during task-specific adaptation~\cite{howard2018universal}. Multimodal learning has been widely studied in intensive care medicine. SMART models irregular multimodal data for patient status prediction~\cite{yu2024smartpretrainedmissingawaremodel}. Other studies combine electronic health records, physiological signals, or clinical notes for outcome prediction~\cite{zhang2023improvingmedicalpredictionsirregular,SUN2024111160,khadanga2020usingclinicalnotestime}. In post-cardiac-arrest cohorts, multimodal models have also been developed for functional outcome prediction~\cite{10364030,krones2024multimodaldeeplearningapproach}. Maschke et al.~\cite{maschke2023functional} used the I-CARE database and combined clinical, ECG, and EEG-derived features. Their analysis emphasized EEG spectral properties, signal complexity, and functional connectivity. These studies demonstrate the prognostic value of multimodal physiological signals. However, they rely heavily on EEG or multiple biosignal modalities. NeuroECG instead focuses on bedside ECG and static clinical covariates. It adapts a pretrained ECG foundation model and aggregates segment embeddings into patient-level representations.

\begin{figure*}[t]
    \centering
    \includegraphics[
        width=1\textwidth,
    ]{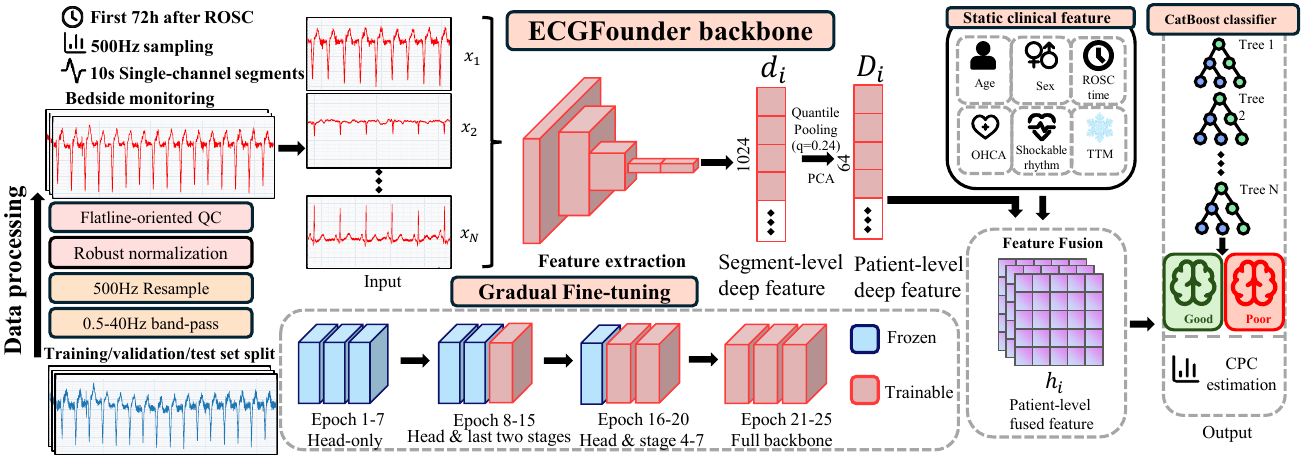}
    \caption{
    Overview of the proposed ECGFounder-based neurological prognostication framework. 
Bedside monitoring ECG recordings within the first 72 hours after ROSC are divided into 10-second single-channel segments and encoded by the ECGFounder backbone. 
Segment-level deep embeddings are aggregated into a patient-level representation using quantile pooling and further compressed by PCA. 
The compact deep ECG representation is combined with static clinical covariates for patient-level CatBoost prediction.
    }
    \label{fig:framework}
\end{figure*}

\section{Methodology}
\subsection{Overview}
Fig.~\ref{fig:framework} illustrates the overall workflow of the proposed NeuroECG framework. The framework contains a primary deep ECG representation pathway and a patient-level prediction module with static clinical covariate integration. For each patient, bedside monitoring ECG recordings within the first 72 hours after ROSC are divided into a set of 10-second single-channel ECG segments:
\begin{equation}
X_i = \{x_{i,k}\}_{k=1}^{N_i},
\end{equation}
where $x_{i,k}$ denotes the $k$-th valid ECG segment of patient $i$.
Each segment is encoded by ECGFounder as
\begin{equation}
d_{i,k}=E_\theta(x_{i,k}),
\end{equation}
where $d_{i,k}\in\mathbb{R}^{1024}$.
The segment embeddings are aggregated by feature-wise quantile pooling
($q=0.24$), compressed to 64 dimensions by PCA, concatenated with
static clinical covariates, and fed to CatBoost for patient-level prediction.

\subsection{ECG Preprocessing and Segment Construction}\label{subsec:preprocessing}
The I-CARE database provides continuous ICU bedside ECG recordings with five channel labels: ECG, ECG1, ECG2, ECGL, and ECGR. We used a deterministic channel selection rule to handle heterogeneous channel naming. We first selected the ECG1/ECG2 pair, followed by the ECGL/ECGR pair. If neither pair was available, one channel was selected in the order ECG, ECG1, ECG2, ECGL, and ECGR. Each valid ECG channel was treated as an independent single-channel sample. When two channels were valid, both were retained as separate samples and jointly contributed to patient-level aggregation. When only one channel was valid, only that channel was retained.

Each selected ECG signal was bandpass-filtered between 0.5 and 40~Hz and resampled to 500~Hz. The signal was then median-centered and robustly scaled by its 95th percentile. For a filtered signal $s(t)$, the normalized signal $\tilde{s}(t)$ was defined as:
\begin{equation}
\tilde{s}(t)=
\mathrm{clip}\left(
\frac{s(t)-\mathrm{median}(s)}
{P_{95}(|s-\mathrm{median}(s)|)+\epsilon},
-5,5
\right),
\end{equation}
where $P_{95}(\cdot)$ denotes the 95th percentile and $\epsilon$ is a small constant for numerical stability.

The normalized signal was divided into 10-second segments. Since the sampling rate was 500~Hz, each segment contained 5,000 samples. A 20-second stride was used to reduce highly repetitive adjacent segments while preserving longitudinal coverage over the first 72 hours after ROSC.

A segment-level flatline-oriented quality-control mask was then applied. Segments with very small amplitude variation were regarded as electrode-off or flatline noise. The validity indicator for segment $x_{i,k}$ was defined as:
\begin{equation}
m_{i,k}=\mathbb{I}\left(\mathrm{std}(x_{i,k}) \geq 0.05\right),
\end{equation}
where $m_{i,k}=1$ indicates a valid ECG segment. Only valid segments were retained for downstream ECGFounder encoding. These retained segments form the patient-level segment set $X_i$ used in Eq.~(1).

To prevent data leakage, all splits were performed at the patient level on the ECG-available cohort. We used joint stratification based on treatment hospital sites (A--F) and clinical outcome labels. Following the I-CARE outcome definition, good neurological outcome was defined as CPC 1--2, whereas poor outcome was defined as CPC 3--5. Poor outcome was treated as the positive class ($y=1$).

\subsection{ECGFounder-Based Deep ECG Representation}
The deep ECG encoder uses the single-lead ECGFounder checkpoint. Its Net1D backbone has a RegNet-style hierarchy of residual bottleneck stages with Swish activations, temporal convolutions, shortcut connections, and Squeeze-and-Excitation modules. This pretrained architecture supplies morphological features from each 10-second single-channel segment. The stage design is inherited from ECGFounder; our adaptation concerns its prediction head and fine-tuning schedule.

The checkpoint was originally trained for ECG diagnosis rather than post-arrest outcome prediction. Consequently, we first train a new outcome head while holding the morphological feature extractor fixed, then progressively allow deeper backbone layers to adapt. This separates the transfer of general ECG structure from task-specific adjustment on a relatively small cohort. All fine-tuning uses the training patients only; the validation patients determine early stopping, and the test patients remain unseen until final evaluation.

In our implementation, the original diagnostic classification head of ECGFounder is removed. To compress the final-stage multidimensional feature map $H^{(S)}$ into a compact representation, a global average pooling (GAP) operator is applied across the temporal dimension to extract the segment-level deep ECG embedding $d_{i,k} \in \mathbb{R}^{1024}$:
\begin{equation}
d_{i,k} = \text{GAP}(H^{(S)})
\end{equation}
During task-specific fine-tuning on the I-CARE cohort, a lightweight binary classification head is attached:
\begin{equation}
\hat{y}_{i,k} = \sigma(W d_{i,k} + b)
\end{equation}
where $\hat{y}_{i,k}$ is the predicted probability of poor neurological outcome for segment $x_{i,k}$. Each segment inherits the patient-level neurological outcome label during fine-tuning. The classification head is optimized using binary cross-entropy loss.

To mitigate overfitting and preserve pretrained ECG morphology, a gradual unfreezing strategy with discriminative learning rates is implemented. The fine-tuning schedule is structured over 25 epochs as follows:

\begin{itemize}
    \item \textbf{Epochs 1--7 (Head-only Phase)}: The entire ECGFounder backbone is frozen. Only the newly added binary classification head is optimized to align the classification objective.
    \item \textbf{Epochs 8--15 (Last-two-stages Unfreezing)}: The last two stages of the ECGFounder backbone, representing stages 6 and 7, are unfrozen. These stages are trained alongside the classification head.
    \item \textbf{Epochs 16--20 (Stages 4--7 Unfreezing)}: Deeper stages of the backbone, representing stages 4 to 7, are unfrozen. This fine-tuning step enables the network to adapt high-level temporal patterns.
    \item \textbf{Epochs 21--25 (Full Backbone Fine-tuning)}: The entire ECGFounder backbone is unfrozen. It is fine-tuned with a smaller learning rate to ensure smooth gradient updates and avoid destroying pretrained features.
\end{itemize}
Phase-specific learning rates were used during fine-tuning. The rates were \(1\times10^{-4}\) for the head, \(1\times10^{-5}\) for the last two stages, \(5\times10^{-6}\) for the additionally unfrozen stages, and \(1\times10^{-6}\) for the remaining backbone.

The progressively smaller rates limit updates to early layers, which carry general waveform features, while permitting stronger task-specific adjustment in the head. The schedule was assessed against head-only training and immediate full unfreezing in the ablation study; no test-set result was used to set the schedule.

During segment-level fine-tuning, we used per-patient capped dynamic sampling. This strategy limits the maximum number of segment-level training samples contributed by each patient, thereby reducing the dominance of patients with substantially longer ECG recordings. At the beginning of each epoch, segments were sampled within each patient. For patient $i$ with $N_i$ valid segments, the epoch-level training subset was defined as:
\begin{equation}
\widetilde{X}_i^{(e)} =
\begin{cases}
X_i, & N_i \leq 256, \\
\mathrm{Sample}_{\mathrm{w/o\ replacement}}(X_i, 256), & N_i > 256,
\end{cases}
\end{equation}
where $e$ denotes the training epoch. Patients with 256 or fewer valid segments contributed all available segments. Patients with more than 256 valid segments contributed 256 segments sampled without replacement. This sampling was repeated at every epoch. The random seed was set as the base seed plus the epoch index. Thus, the same base seed produces the same sampled segments at each epoch.
\begin{table*}[hbtp]
\centering
\caption{Performance comparison of different ECG backbones under the same ECG-only and PCA-compressed evaluation settings. $\dagger$ denotes backbones initialized with pretrained weights.}
\label{tab:backbone_comparison}
\resizebox{\textwidth}{!}{%
\begin{tabular}{lcccccc}
\toprule
\multirow{2}{*}{\textbf{Backbone}} & \multicolumn{3}{c}{\textbf{Deep-only (Raw)}} & \multicolumn{3}{c}{\textbf{Deep PCA-only}} \\
\cmidrule(r){2-4} \cmidrule(l){5-7}
& \textbf{test AUROC} & \textbf{test AUPRC} & \textbf{test CPC-MAE} & \textbf{test AUROC} & \textbf{test AUPRC} & \textbf{test CPC-MAE} \\
\midrule
ConvNeXt1D & $0.5714 \pm 0.0200$ & $0.7382 \pm 0.0238$ & $1.6865 \pm 0.0105$ & $0.4884 \pm 0.0387$ & $0.6734 \pm 0.0378$ & $1.7156 \pm 0.0097$ \\
SE-ResNet  & $0.5664 \pm 0.0228$ & $0.7491 \pm 0.0231$ & $1.6804 \pm 0.0177$ & $0.5215 \pm 0.0573$ & $0.6987 \pm 0.0469$ & $1.6952 \pm 0.0125$ \\
ECG-JEPA$^\dagger$   & $0.5828 \pm 0.0426$ & $0.7567 \pm 0.0323$ & $1.6787 \pm 0.0125$ & $0.6510 \pm 0.0911$ & $0.7857 \pm 0.0637$ & $1.6832 \pm 0.0272$ \\
ECG-FM$^\dagger$     & $0.6649 \pm 0.0365$ & $0.8249 \pm 0.0292$ & $1.6512 \pm 0.0213$ & $0.6950 \pm 0.0136$ & $0.7976 \pm 0.0123$ & $1.6254 \pm 0.0232$ \\
ST-MEM$^\dagger$     & $0.5659 \pm 0.0763$ & $0.7320 \pm 0.0651$ & $1.7045 \pm 0.0097$ & $0.6172 \pm 0.0353$ & $0.7666 \pm 0.0299$ & $1.6704 \pm 0.0125$ \\
ECGFounder$^\dagger$ & $\mathbf{0.7333 \pm 0.0187}$ & $\mathbf{0.8540 \pm 0.0163}$ & $\mathbf{1.5655 \pm 0.0172}$ & $\mathbf{0.7097 \pm 0.0108}$ & $\mathbf{0.8011 \pm 0.0231}$ & $\mathbf{1.5890 \pm 0.0202}$ \\
\bottomrule
\end{tabular}%
}
\end{table*}
\begin{table*}[hbtp]
\centering
\caption{Downstream outcome classification (AUROC, AUPRC, F1-score, Recall) and cerebral performance category regression (CPC-MAE) results of different backbones under the proposed NeuroECG configuration (Deep PCA + Static).}
\label{tab:backbone_fusion}
\begin{tabular}{lccccc}
\toprule
\multirow{2}{*}{\textbf{Backbone}} & \multicolumn{5}{c}{\textbf{Deep PCA + Static}} \\
\cmidrule(r){2-6}
& \textbf{test AUROC} & \textbf{test AUPRC} & \textbf{test F1-score} & \textbf{test Recall} & \textbf{test CPC-MAE} \\

\midrule
SE-ResNet  & $0.7134 \pm 0.0220$ & $0.8014 \pm 0.0195$ & $0.7989 \pm 0.0242$ & $0.8143 \pm 0.0315$ & $1.6204 \pm 0.0381$ \\
ConvNeXt1D & $0.7356 \pm 0.0225$ & $0.8464 \pm 0.0147$ & $0.7805 \pm 0.0396$ & $0.7810 \pm 0.0259$ & $1.6289 \pm 0.0383$ \\
ECG-FM$^\dagger$     & $0.7655 \pm 0.0199$ & $0.8601 \pm 0.0213$ & $0.8037 \pm 0.0153$ & $0.8095 \pm 0.0240$ & $1.5508 \pm 0.0553$ \\
ECG-JEPA$^\dagger$   & $0.7660 \pm 0.0121$ & $0.8619 \pm 0.0073$ & $0.7960 \pm 0.0171$ & $0.8190 \pm 0.0358$ & $1.5923 \pm 0.0515$ \\
ST-MEM$^\dagger$     & $0.7680 \pm 0.0139$ & $0.8758 \pm 0.0167$ & $0.8175 \pm 0.0238$ & $0.8333 \pm 0.0336$ & $1.5537 \pm 0.0552$ \\
ECGFounder$^\dagger$ & $\mathbf{0.8077 \pm 0.0170}$ & $\mathbf{0.8970 \pm 0.0082}$ & $\mathbf{0.8218 \pm 0.0181}$ & $\mathbf{0.8524 \pm 0.0261}$ & $\mathbf{1.5156 \pm 0.0364}$ \\
\bottomrule
\end{tabular}
\end{table*}
This dynamic sampling strategy was used only during segment-level fine-tuning. After fine-tuning, patient-level feature extraction was performed on the available valid ECG segments. The segment-level classification head was discarded. The adapted ECGFounder backbone was frozen and retained as a deep feature extractor. Because the prediction target is defined at the patient level, segment-level embeddings must be aggregated into a unified patient-level representation. For patient $i$, the segment embeddings $\{d_{i,k}\}_{k=1}^{N_i}$ are aggregated via feature-wise quantile pooling. This pooling step is applied independently to each of the 1024 dimensions:
\begin{equation}
z_i = Q_q(\{d_{i,k}\}_{k=1}^{N_i}), \quad q = 0.24
\end{equation}
where $Q_q(\cdot)$ extracts the $q$-quantile. Then, the aggregated patient-level representation $z_i$ is compressed using principal component analysis (PCA). For a PCA dimension $m$, the compressed patient-level deep feature is defined as:
\begin{equation}
D_i^{(m)} = U_m^T \frac{z_i - \mu_{\text{tr}}}{\sigma_{\text{tr}}},
\end{equation}
where $\mu_{\text{tr}}$ and $\sigma_{\text{tr}}$ are estimated only from the training set. $U_m$ contains the first $m$ principal directions fitted on the training set. In the main NeuroECG configuration, we use \(m=64\). This dimension was selected based on preliminary validation experiments. It substantially reduces the dimensional gap between the 1024-dimensional ECG representation and the six static covariates while avoiding overly aggressive compression. PCA therefore serves as a compact bottleneck for patient-level fusion.

For validation and test patients, the training-set scaling parameters and PCA directions are applied without refitting. The quantile is computed across all valid segments available for each patient, including both selected channels when present. Thus, a patient contributes one fixed-length vector to the downstream model regardless of monitoring duration, and all patient-level evaluation uses one prediction per patient.

\subsection{Patient-Level Fusion and Prediction}
The proposed prediction model combines the compact patient-level deep ECG representation $D_i \in \mathbb{R}^{64}$ with six static clinical covariates $c_i \in \mathbb{R}^6$. The static branch used six patient-level clinical covariates provided in the I-CARE data used in this study: age, sex, time to ROSC, out-of-hospital cardiac arrest, shockable rhythm, and TTM temperature. The patient-level fused feature vector $h_i$ is constructed via concatenation:
\begin{equation}
h_i = \text{Concat}[D_i, c_i]
\end{equation}
For downstream prediction, we employ a CatBoost classifier to predict the probability of poor neurological outcome:
\begin{equation}
\hat{p}_i = \Phi_{\text{cat}}(h_i)
\end{equation}
A separate CatBoost regressor was used for secondary CPC score estimation. CPC was treated as an ordered numerical target, and performance was evaluated using MAE.

\section{Experiments} \label{sec:experiments}

\subsection{Experimental Setup}
Of the 607 I-CARE patients, 412 had usable ECG recordings and were included in this study. The cohort was stratified by hospital site and outcome and split into training, validation, and test sets of 288, 61, and 63 patients, respectively. The split was fixed with seed 42.

Stage 1 fine-tuning used segment-level validation BCE for early stopping with a patience of 6 epochs. Per-patient capped sampling retained at most 256 valid segments per patient in each epoch. The quantile threshold q was selected by five-fold cross-validation on the training cohort. Within each fold, ECGFounder adaptation, standardization, PCA, and CatBoost were fitted using only the fold-specific training patients. The held-out fold was used only for validation. Final evaluation was performed at the patient level.

CatBoost used 100 iterations, a depth of 4, and a learning rate of 0.03. Class weights were 1.7 for the minority good-outcome class and 1.0 for the poor-outcome class. Downstream models used validation-based early stopping and were trained with seeds 42--46. Results are reported as mean $\pm$ standard deviation. Patient-level paired bootstrap was used to estimate 95\% confidence intervals for AUROC and the AUROC difference between NeuroECG and the static-only baseline. All backbone comparisons used the same patient split, preprocessing, pooling, PCA dimension, and downstream settings. Pretrained models are marked with $\dagger$; SE-ResNet and ConvNeXt1D were trained from scratch.

\subsection{Experimental Result}

Table~\ref{tab:backbone_comparison} compares the ECG-only backbones using raw and PCA-compressed patient-level representations. With raw embeddings, ECGFounder had the highest mean test AUROC (0.7333) and AUPRC (0.8540), and the lowest CPC-MAE (1.5655). Its AUROC was lower after PCA compression (0.7097), although it remained the highest among the PCA-only backbones. PCA yielded higher ECG-only AUROC point estimates for ECG-JEPA, ECG-FM, and ST-MEM, but not for ECGFounder. Thus, PCA is used to reduce feature dimensionality before fusion; these ECG-only results do not show that it generally improves discrimination.

Table~\ref{tab:backbone_fusion} reports the five outcome metrics after combining each backbone's PCA-compressed ECG representation with six static clinical covariates. ECGFounder had the highest mean test AUROC (0.8077), AUPRC (0.8970), F1-score (0.8218), and recall (0.8524), as well as the lowest CPC-MAE (1.5156). The static-only baseline had an AUROC of 0.7689 and an AUPRC of 0.8574 (Table~\ref{tab:feature_ablation}). The fused model's corresponding point estimates were higher by 0.0388 and 0.0396.

Using seed-ensemble patient predictions, paired bootstrap analysis yielded an AUROC of 0.8175 (95\% CI: 0.6979--0.9194) for NeuroECG and 0.7710 (95\% CI: 0.6304--0.8913) for the static-only baseline. The paired AUROC difference was 0.0465 (95\% CI: -0.0293--0.1263). Because this interval includes zero, these data suggest a possible benefit from ECG fusion but do not establish a clear AUROC gain. The ensemble estimates differ from the five-seed means reported in the tables because they are calculated from ensemble predictions.

\subsection{Ablation Study}
Table~\ref{tab:feature_ablation} separates the contribution of ECG features, static covariates, and PCA compression. Static + Deep improved AUROC over static-only from $0.7689$ to $0.7931$. Adding PCA further increased AUROC to $0.8077$ while reducing the fused dimension from 1030 to 70. PCA did not improve standalone ECG discrimination ($0.7333$ versus $0.7097$ AUROC), so its observed benefit was specific to the fusion setting.

\begin{table}[hbtp]
\centering
\caption{Feature ablation on the test set (mean $\pm$ standard deviation).}
\label{tab:feature_ablation}
\resizebox{\columnwidth}{!}{%
\begin{tabular}{lcccc}
\toprule
\textbf{Features} & \textbf{Dim.} & \textbf{AUROC} & \textbf{AUPRC} & \textbf{CPC-MAE} \\
\midrule
Static only & 6 & $0.7689 \pm 0.0034$ & $0.8574 \pm 0.0023$ & $1.5762 \pm 0.0274$ \\
Deep only & 1024 & $0.7333 \pm 0.0187$ & $0.8540 \pm 0.0163$ & $1.5655 \pm 0.0172$ \\
Deep PCA-only & 64 & $0.7097 \pm 0.0108$ & $0.8011 \pm 0.0231$ & $1.5890 \pm 0.0202$ \\
Static + Deep & 1030 & $0.7931 \pm 0.0040$ & $0.8866 \pm 0.0033$ & $1.5494 \pm 0.0026$ \\
Static + Deep PCA & 70 & $\mathbf{0.8077 \pm 0.0170}$ & $\mathbf{0.8970 \pm 0.0082}$ & $\mathbf{1.5156 \pm 0.0364}$ \\
\bottomrule
\end{tabular}%
}
\end{table}

Pooling and backbone adaptation were tested under otherwise identical settings (Table~\ref{tab:strategy_ablation}). Quantile pooling outperformed mean and max pooling in the fused model. Five-fold cross-validation on the training cohort selected $q=0.24$, with validation AUROC $0.8093$; the test set was not used to choose $q$. Gradual unfreezing gave the highest ECG-only AUROC among the three adaptation schedules.

Mean pooling treats persistent and transient segment responses similarly, whereas max pooling emphasizes the strongest response in each embedding dimension. Feature-wise low-quantile pooling instead summarizes a lower portion of each dimension's distribution over the patient's valid segments. The comparisons support this choice for the present cohort but do not identify a physiological interpretation for any single latent dimension. Likewise, the weaker result from immediate full unfreezing is consistent with unstable adaptation under limited patient-level data, although this ablation alone does not establish the mechanism.

\begin{table}[hbtp]
\centering
\caption{Key pooling and fine-tuning ablations on the test set. Pooling uses Static + Deep PCA; fine-tuning uses ECG-only features.}
\label{tab:strategy_ablation}
\resizebox{\columnwidth}{!}{%
\begin{tabular}{lcc}
\toprule
\textbf{Configuration} & \textbf{AUROC} & \textbf{AUPRC} \\
\midrule
Mean pooling & $0.7730 \pm 0.0233$ & $0.8615 \pm 0.0285$ \\
Max pooling & $0.7662 \pm 0.0098$ & $0.8607 \pm 0.0099$ \\
Quantile pooling ($q=0.24$) & $\mathbf{0.8077 \pm 0.0170}$ & $\mathbf{0.8970 \pm 0.0082}$ \\
\midrule
Head-only fine-tuning & $0.6968 \pm 0.0181$ & $0.8499 \pm 0.0125$ \\
Immediate full unfreezing & $0.6213 \pm 0.0146$ & $0.7759 \pm 0.0135$ \\
Gradual unfreezing & $\mathbf{0.7333 \pm 0.0187}$ & $\mathbf{0.8540 \pm 0.0163}$ \\
\bottomrule
\end{tabular}
}
\end{table}

\subsection{Model Interpretability}
Test-set SHAP analysis~\cite{lundberg2017unified} ranked shockable rhythm (25.4\%) and age (9.0\%) as the leading individual features. Deep PCA components 49, 32, and 25 contributed 4.9\%, 4.8\%, and 4.3\%, respectively. Grouped importance was 41.96\% for static covariates and 58.04\% for deep ECG components; because the groups contain different numbers of features, these totals should not be interpreted as a direct per-feature comparison.

\begin{figure}[hbtp]
\centering
\includegraphics[width=\columnwidth,trim=52bp 0bp 52bp 0bp,clip]{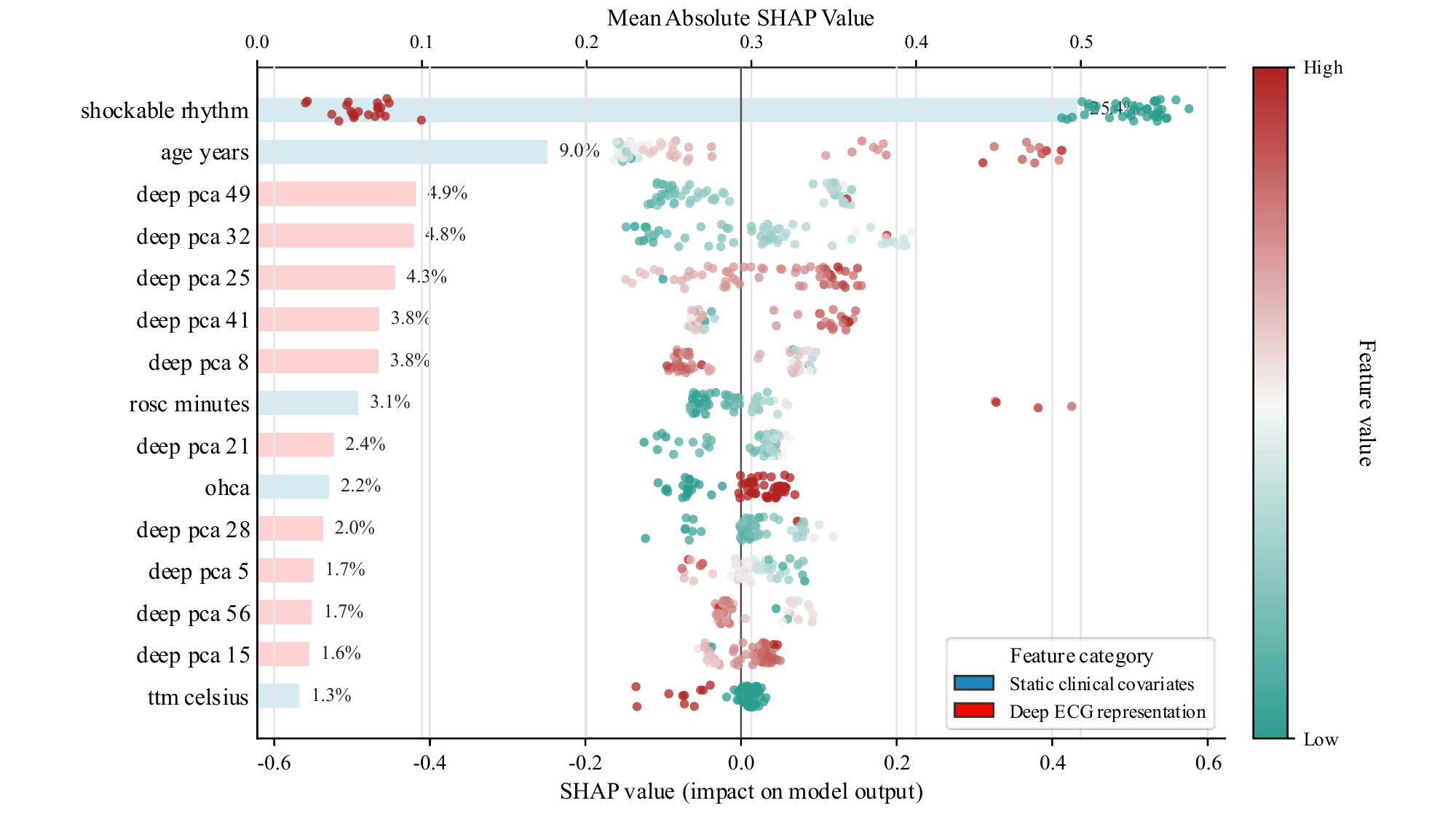}
\caption{Test-set SHAP summary for the fused NeuroECG model. The 15 leading features include static clinical variables and PCA-compressed ECG components. Positive SHAP values indicate a larger predicted probability of poor neurological outcome.}
\label{fig:shap_importance}
\end{figure}

\section{Limitations}

First, although I-CARE is multicenter, all experiments used this database and no independent external validation was performed. To our knowledge, no other public post-cardiac-arrest dataset provides comparable bedside ECG recordings with neurological outcome labels; validation on independent institutional cohorts is therefore needed.

Second, usable ECG recordings were unavailable in the released I-CARE data for 195 of 607 patients. The 412-patient analytic cohort thus reflects ECG availability in the source dataset rather than additional patient-level screening by the investigators. This limited coverage may affect generalizability and warrants evaluation in cohorts with more complete ECG data.

Third, segment-level fine-tuning used patient-level outcome labels, introducing weak supervision because individual segments lack time-resolved annotations. An exploratory multiple-instance learning (MIL) approach to patient-level training performed less well in this cohort. Future work should examine improved patient-level or temporally resolved supervision.

Finally, the PCA-compressed ECG features are latent and cannot yet be linked directly to clinically interpretable waveform patterns. Establishing these links would strengthen the physiological interpretation of the model.

\section{Conclusion}
In this study, we proposed NeuroECG, a deep ECG representation framework for EEG-free auxiliary neurological prognostication after cardiac arrest. The framework fine-tunes a pretrained ECG foundation model using a gradual unfreezing strategy. It aggregates segment embeddings via quantile pooling and compresses them using principal component analysis. Experiments on the multicenter I-CARE database show that ECGFounder achieves the best performance among the evaluated ECG-only backbone baselines. The proposed multimodal model further combines deep ECG representations with static clinical covariates. It improves over clinical covariates alone and achieves the best overall performance in our experiments. These results suggest that bedside ECG contains useful auxiliary prognostic information. These findings support further investigation of deep bedside ECG representations as an auxiliary EEG-independent source of prognostic information.

\section*{Acknowledgment}
This work was supported by “the Fundamental Research Funds for the Central Universities” under Grant X202610701781.

\bibliographystyle{IEEEtran}
\bibliography{ECGfounder}
\end{document}